\documentclass{article}
\usepackage{framed,multirow}

\usepackage{amssymb}
\usepackage{amsmath}
\usepackage{latexsym}

\usepackage{url}
\usepackage{xcolor}
\definecolor{newcolor}{rgb}{.8,.349,.1}

\usepackage[ruled,vlined]{algorithm2e}

\usepackage{graphicx}
\usepackage{subcaption}
\usepackage{mwe}
\usepackage{color,soul}

\title{Instance-based learning using the Half-Space Proximal Graph\footnote{Under review in Pattern Recognition Letters}}

\author{Ariana Talamantes \\ {\tt ariana@cicese.edu.mx }\and Edgar Chavez \\  {\tt elchavez@cicese.mx}}

\begin{document}

\maketitle
\begin{abstract}
The primary example of instance-based learning is the $k$-nearest neighbor rule (kNN), praised for its simplicity and the capacity to adapt to new unseen data and toss away old data. The main disadvantages often mentioned are the classification complexity, which is $O(n)$, and the estimation of the parameter $k$, the number of nearest neighbors to be used. The use of indexes at classification time lifts the former disadvantage, while there is no conclusive method for the latter. 

This paper presents a parameter-free instance-based learning algorithm using the {\em Half-Space Proximal} (HSP) graph. The HSP neighbors simultaneously possess proximity and variety concerning the center node. To classify a given query, we compute its HSP neighbors and apply a simple majority rule over them. In our experiments, the resulting classifier bettered $KNN$ for any $k$ in a battery of datasets. This improvement sticks even when applying weighted majority rules to both kNN and HSP classifiers.

Surprisingly, when using a probabilistic index to approximate the HSP graph and consequently speeding-up the classification task, our method could {\em improve} its accuracy in stark contrast with the kNN classifier, which worsens with a probabilistic index. 
\end{abstract}

\section{Introduction}

One of the most popular classifiers is k-Nearest Neighbors (kNN), which was rated as one of the top 10 algorithms in data mining \cite{Settouti2016, Wu2008}. Trivially implemented, kNN is popular because of its simplicity. There is no training stage, and as soon as the data is acquired, the algorithm is ready to make predictions. Moreover, it works without prior knowledge of the data distribution. \par

A vanilla implementation of the kNN classifier consists of computing the distance between the query and every training sample to obtain a neighborhood of the closest $k$ samples to the query and assigning the majority's label in that neighborhood. The kNN classifier's performance depends crucially on the neighborhood's size determined by the value of $k$ \cite{Gou2019, Zhang2018} and the distance function used to measure the similarity between two specimens \cite{Weinberger2005, Xu2013}. 

Choosing the optimal parameter $k$ for kNN is a challenging task. A large  $k$ produce a neighborhood robust to noise, but it may include too many neighbors from other classes \cite{Gallego2018}, while a small value of \textit{k} often leads to an overfitted decision boundary resulting in high noise sensitivity. Ideally, \textit{k} should be adapted to the every dataset in particular. 
%A trivial solution would be to try different values of $k$ and then choose the one that works best for the particular problem, but doing so is expensive and requires prior knowledge or learning the data distribution. \par

% more about the problem of k 

On the other hand, regarding complexity or speed, a brute force algorithm for finding the $k$-nearest neighbors has linear complexity, $O(n)$ with $n$ the database size.  This complexity does not scale for large problems. Unlike databases made up of simple attribute data, recent data tends to be large and complex. For example, in multimedia data, the standard approach is to search not at the level of actual multimedia objects but instead using the so-called deep-features extracted from these objects \cite{Zezula2005}. This technique produces high-dimensional vectors that are difficult to index, with provable hardness complexity\cite{r2018STOC}.  
The difficulty of finding the $k$ closest neighbors in large, high-dimensional databases has prompted approximation algorithms for the search for similarity (ANN). An ever-growing amount of research in ANN search aims at high accuracy and low computational complexity algorithms. These methods are typically used as an offline stage in kNN to accelerate the classification tasks. \par
For high dimensional spaces, graph-based ANN takes the lead in algorithm usage. One of the most efficient graphs for the nearest neighbor search is the Small World graph (SW-graph), proposed in \cite{Malkov2014} as NSW graph. Each insertion finds its approximate neighbors by a greedy search in the partial graph built so far. HNSW \cite{Malkov2020} is an extension of NSW. It incrementally builds a multi-layer structure consisting of a hierarchical set of proximity graphs (layers) for nested subsets of the stored elements. HNSW is one of the most efficient, general-purpose algorithms \cite{Li2020} for this problem. HNSW is the index we selected to speed-up near-neighbor searches in this work. \par

\subsection{Motivation}

Probabilistic indexes, such as HNSW, have lifted one of the main disadvantages of the kNN classifier: classification speed.  On the other hand, deep-features simplifies the design of the distance to compare instances in the classifier. One standing problem in kNN classification is the selection of the parameter $k$. Researchers have proposed many methods to discover a {\em good} value for $k$ (discussed with more detail later, in the next section) in a quest to lift this last limitation for a proper black-box classification algorithm. This paper proposes a more general approach; designing an algorithm that naturally chooses a query neighborhood without parameters. This neighborhood should contain objects near the query while at the same time providing {\em geometric diversity} within the database. In other words, we want to eliminate redundancy in the neighborhood.

\subsection{Contribution}

We base our proposal on the Half-Space Proximal (HSP) graph introduced in \cite{Chavez2006}. This graph extracts a low-degree spanner of the complete graph. Each node is associated with its nearest neighbor, clearing from the complete graph all redundant nodes in the nearest neighbor's direction by using a half-space hyperplane. We repeat with the remaining nodes until clearing all of them. We will discuss this in more detail in section \ref{sec:HSP}. We fixed our attention just in the neighborhood of the query instead of the entire graph.

In this paper, we show that a majority classifier using the HSP neighborhood of the query systematically outperforms the kNN classifier for {\em any} $k$. Computing the HSP neighborhood of the query is as fast as computing the $k$-nearest neighbors, and it admits speeding-up using an index. Moreover, when using an index, the classification precision {\em increases}, unlike the kNN classifier. We tested our claims with a realistic experimental setup with a well-known benchmark. 

%%%%%%%%%%%%%%

\section{Related Work}

Research in kNN performance is abundant, with classical and recent approaches\cite{Gou2019}, with many variations of the classifier \cite{Wu2008, Shi2018}. The efforts focus mostly on solving one or more of the issues present in kNN. \par
In the vanilla kNN algorithm, the distance function used is Euclidian, with the same weight to all features, yielding inaccurate results when irrelevant attributes are present, as in high-dimensional data \cite{Jiang2007}. The approaches for solving this problem include assigning different weights to each feature \cite{Biswas2018} or eliminating the least relevant features. Some other methods include assigning different weights to each neighbor, with the idea that closer neighbors should contribute more for assigning the class label to the query \cite{Tang2020}. Other approaches include the design of distance functions like  Mahalanobis \cite{Xiang2008,Gautheron2020}, adaptive Euclidian \cite{Wang2007}, or the Value Difference Metric (VDM) \cite{Li2011,Li2014}. \par

In choosing the optimal $k$, the vanilla approach uses a fixed value of $k$ for every test sample. A popular choice is to use  $k=\sqrt{n}$, proposed in \cite{Lall1996}. Another method, proposed in \cite{Zhu2011}, is  tenfold cross-validation to find the optimal value for $k$. However, in \cite{Zhang2017}  they show that a fixed value leads to a low prediction rate since it does not consider the distribution of the data. Alternatively, recent efforts focus on setting a different $k$ value for each test sample, giving better results \cite{Liu2010}. More efforts in this direction include evolutionary computation techniques \cite{Biswas2018}, probabilistic methods \cite{Ghosh2006}, and linear modeling \cite{Zhang2017}, among others. \par

In general, the idea behind these methods is to search for the optimal $k$ values and then perform a traditional kNN classification. The methods that use this approach require additional processing time during the classification, which increases the algorithm's overall complexity. According to \cite{ktree2018} these techniques have a time complexity of at least \(O(n^2)\) during the classification time, which is not suitable for large data repositories. Zhang et al. \cite{ktree2018} proposed the \textit{kTree} and \textit{k*Tree} methods, which introduce an offline training stage that, in addition to finding a proper size of the neighborhood, also focuses on reducing the online classification time. This approach also has quadratic complexity to find the neighborhood's optimal size,  finding optimal $k$ values in an offline stage. With this modification, instead of having an online complexity of $O(n^2)$ during classification, the task can be achieved with a complexity of $O(log(d)+n)$, where $d$ is the dimensions of the features.

For each of the proposed methods and improvements of kNN, there is an additional time-consuming stage, online or offline, to estimate a proper \textit{k} value of each test sample. These procedures take away the simplicity of kNN, which is one of the characteristics that makes it so popular. Moreover, the accuracy of these methods is limited to the traditional kNN with optimal parameters. \par

\section{HSP graph}
\label{sec:HSP}
The HSP graph is a \textit{local proximity graph} \cite{Jaromczyk1992} that was originally proposed and designed for applications in \textit{ad-hoc networks}. Computationally, these networks are represented by Unit Disk Graphs (UDG), where the nodes represent the network components, called hosts. An edge connects two nodes if the Euclidian distance between the hosts is less than a given unit, where the unit represents the common transmission range of the hosts in the network. An edge indicates the hosts can communicate with each other with a single transmission, called a hop.\par

\begin{figure}[ht]
\centering
\includegraphics[scale=.3]{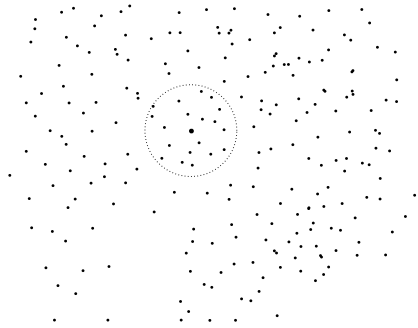}
\caption{Example of the transmission range of a host. In a UDG, all nodes within the range would be connected to the central node.}
\label{fig:hsp range}
\end{figure}

\begin{figure}[ht]
\centering
\includegraphics[scale=.3]{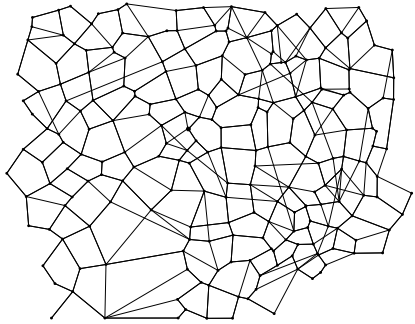}
\caption{Example of an HSP graph.}
\label{fig:hsp example}
\end{figure}

The HSP test determines which neighbors are retained within each node's range for constructing a suitable geometric subgraph of the UDG. The resulting graph referred to as the HSP graph, is a sparse directed or undirected subgraph of the UDG (see Figure \ref{fig:hsp example}). \par 

Extracting a UDG subgraph reduces the complexity of the network,  which is useful in many applications. Some examples include energy-efficient routing and power optimization. These applications often need the resulting graph to have some properties like having a small degree, being planar, or having the minimum spanning tree as its subgraph. \par

The HSP graph is a computationally simple algorithm that has many properties desirable for network applications. An important characteristic is that it uses only local computations for its construction. This property is essential because in \textit{ad-hoc networks} the topology of the whole network is usually not available. Besides, in dynamically changing networks, eventual changes should be detected and fixed without disturbing the entire network. \par 

\subsection{Construction}

\begin{figure}[htp]
    \centering
        \begin{subfigure}[b]{0.13\textwidth}
            \centering
            \includegraphics[width=.7\textwidth]{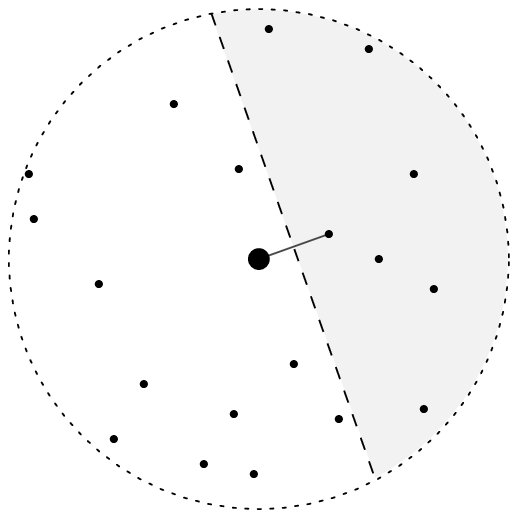}
            \caption{}    
            \label{fig:neighbor 1}
        \end{subfigure}
    \hspace{0em}%
        \begin{subfigure}[b]{0.13\textwidth}  
            \centering 
            \includegraphics[width=.7\textwidth]{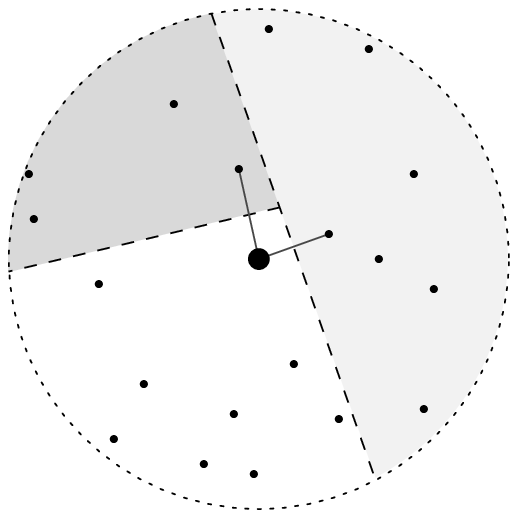}
            \caption{}    
            \label{fig:neighbor 2}
        \end{subfigure}
    \vskip\baselineskip
        \begin{subfigure}[b]{0.13\textwidth}   
            \centering 
            \includegraphics[width=.7\textwidth]{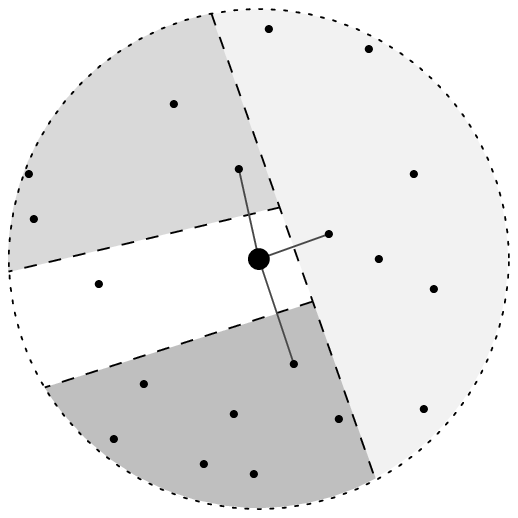}
            \caption{}    
            \label{fig:neighbor 3}
        \end{subfigure}
   \hspace{0em}%
        \begin{subfigure}[b]{0.13\textwidth}   
            \centering 
            \includegraphics[width=.7\textwidth]{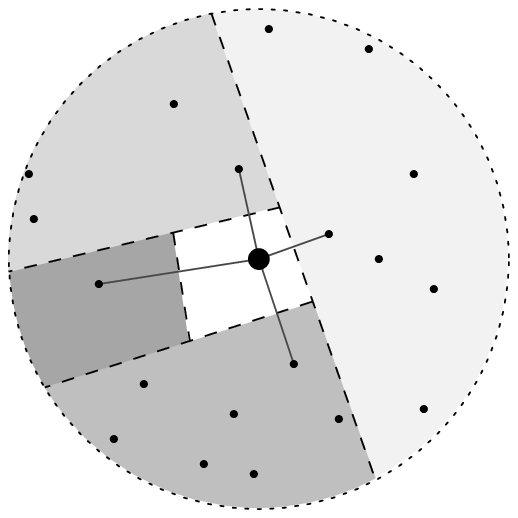}
            \caption{}     
            \label{fig:neighbor 4}
        \end{subfigure}
    \caption{Zooming around the vecinity of a selected node}
    \label{fig:neighbor selection}
\end{figure}

For the construction of the HSP graph, we assume the graph $G=(V,E)$ is a UDG with coordinates $(v_{x}$,$v_{y})$ for each node $v$ in the Euclidian plane, and a unique integer label for each vertex. The algorithm to choose the neighbors for each node to construct the HSP graph is described in Algorithm \ref{algorithm:hsp} and illustrated in Figure \ref{fig:neighbor selection}.\par  

\begin{algorithm}
\SetAlgoLined
\KwIn{a vertex $u$ of a geometric graph and a list $L_{1}$ of edges incident with $u$.}
\KwOut{a list of directed edges $L_{2}$ which are retained for the HSP graph.}

Set the forbidden area $F(u)$ to be $\varnothing$\;

\While{$L_{1}$ is not empty}{
  Remove from $L_{1}$ the shortest edge, say $(u,v)$, (any tie is broken by smaller end-vertex label) and insert in $L_{2}$ the directed edge $(u,v)$ with $u$ being the initial vertex\;
  Add to $F(u)$ the open half-plane determined by the line perpendicular to the edge $(u,v)$ in the middle of the edge and containing the vertex $v$\; 
  Scan the list $L_{1}$ and remove from it any edge whose end-vertex is in $F(u)$\;

 }

 \caption{HSP}
 \label{algorithm:hsp}
\end{algorithm}

From \cite{Chavez2006}, Figure \ref{fig:neighbor selection} illustrates the forbidden half-space represented by a shaded area. Computationally, an edge $(u,z)$ is forbidden by an edge $(u,v)$ when the Euclidian distance from $z$ to $v$ is smaller than the Euclidian distance from $z$ to $u$. Additionally, there is no explicit use of the coordinates, and each node chooses its neighbors without parameters.  \par

\subsection{Properties}
The HSP graph has many desirable properties for ad-hoc networks, it is a $t$-spanner, with $t\ge (2\pi+1)$. The obtained spanner is invariant under similarity transformations and contains the minimum weight spanning tree.\par

The out-degree of the HSP graph depends on the data's intrinsic dimension, and it coincides with the kissing number in that dimension. For example, it is 2, 6, 12, and 24 for dimensions 1, 2, 3, and 4, respectively. In higher dimensions, only upper and lower bounds are known, with few exceptions. The relevant feature for the HSP in classification tasks is that it provides diversity and similarity in each node's neighborhood. The neighborhood of a node comprises the neighbors after eliminating the redundancy between them. Notably, the above can be achieved without tuning parameters. This last property is what makes it unique for classification applications. \par

The HSP graph is fully distributed and computationally simple to construct. The algorithm is executed by each node using only information of their neighborhood. Although initially formulated for two-dimensional data, where the vectors represent the physical coordinates that correspond to geographic locations, the HSP graph's algorithm has no explicit use of the coordinates, which means that it can be generalized to work in any metric space. Therefore, a generalized HSP algorithm can be used for other applications\cite{corral, aguilera}, as our proposal for classification tasks is the case. \par

\section{HSP for classification}

Our proposal is the {\em HSP classifier}, that is, using the neighborhood discovered by the HSP test as the instances to be compared with a query. Our initial proposal assumes each query can choose its neighbors from the entire database (i.e., all the training samples) instead of only within a range determined by a given unit, as it is the UDG case; this corresponds to a UDG with infinite radius.\par 

The selected neighbors have the property of being similar while diverse, which gives a representative vicinity for each query. Furthermore, besides being a computationally simple algorithm, this proposal's main advantage is that there is no parameter $k$ needed for selecting the neighbors; the selection happens naturally. \par

Our proposal solves the problem of choosing the proper $k$ neighbors to do the classification task. Once we obtained the neighborhood, we apply the majority rule over the selected neighbors' labels, as in the vanilla kNN classifier. \par
The proposal could be trivially implemented in any metric space and any dimension. We present the pseudocode in Algorithm \ref{algorithm:hspclassifier}.  \par

\begin{algorithm}
\SetAlgoLined
\KwIn{training samples X and test samples Y}
\KwOut{class labels of Y}
 \For{each $u\in Y$}{
 %$F(u) \leftarrow \varnothing$ \tcp{Set forbidden area}
 $N \leftarrow \varnothing$\;
 $C \leftarrow X$\;
 \While{C is not empty}{
  $v \leftarrow c\in C \mid d(u,c) \leq d(u,c'), \forall c'\in C$\;
  $N.insert(v)$\;
  \For{each $c\in C$}{
  \If{$d(c,u) > d(c,v) $}{
   $C.remove(c)$\;
  }
  }
 }
 
 $label(u) \leftarrow$ most repeated label in $N$\;
 
 }
 \caption{HSP classifier}
 \label{algorithm:hspclassifier}
\end{algorithm}

\section{Experiments}

We performed experiments to assess the performance of the HSP classifier, contrasting it with the kNN. We selected realistic high-dimensional data obtained by performing deep-feature extraction on popular datasets used for image classification tasks. We used the VGG16 model weights pre-trained with ImageNet\cite{Deng2009}. Researchers often call the above process {\em deep-features} or knowledge transfer. In this case, we feed the  VGG16 model with the corresponding image and output the last layer just before classification. The above is a widely used technique for image classification tasks \cite{Gallego2018,Wang2019}. \par

We selected five datasets with different characteristics; those datasets are among the most popular for benchmarking classification algorithms. In table \ref{table:datasets} we describe each of the selected datasets for the experiments.

\begin{table}[!ht]
\centering
\caption{Description of the datasets}
\begin{tabular}{|p{2.9cm}|p{1.4cm}|p{1.4cm}|p{1.4cm}|}
\hline
Name & \# samples & \# features & \# classes  \\
\hline
MNIST \cite{LeCun2010} & 70000 & 512 & 10 \\
Fasion MNIST \cite{Xiao2017} & 70000 & 512 & 10 \\
CIFAR10 \cite{Krizhevsky2009} & 60000 & 512 & 10 \\
CIFAR100 \cite{Krizhevsky2009} & 60000 & 512 & 100 \\
Mini-ImageNet \cite{Vinyals2016} & 60000 & 2048 & 100 \\
\hline
\end{tabular}
\label{table:datasets}
\end{table}

The datasets have an increasing difficulty of classification, respectively MNIST, Fashion MNIST, CIFAR 10, Mini-ImageNet, and CIFAR 100. Each dataset has a dimension that depends on the deep-feature extraction and the original size of the images. Both CIFAR10 and CIFAR100 have an original size of (32x32). MNIST and Fashion MNIST have an original size of (28x28) but were converted to (32x32) since the model accepts this minimum size. Thus, the size of the feature vectors for these four datasets is the same. \par
In the case of mini-Imagenet, the image size is (84x84). This dataset is interesting since it is a subset of ImageNet. The complexity of Mini-ImageNet is relatively high but requires fewer resources than the full ImageNet dataset. \par
 
In each of the datasets, we selected a random sample of 1000 images for testing. Since our focus is on designing a parameter-less instance-based classifier, we only used the Euclidean distance to measure the similarity between samples. \par

\begin{figure}[!ht]
\centering
\includegraphics{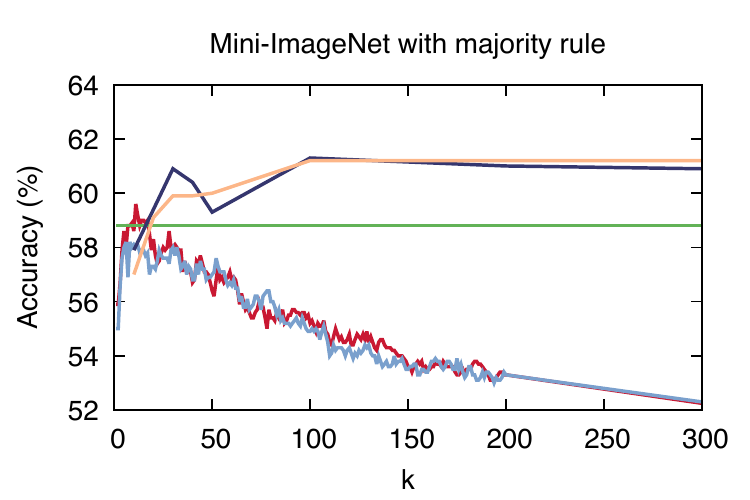}
\caption{Mini-ImageNet - VGG16}
\label{fig:mini:maj}
\end{figure}

\begin{figure}[!ht]
\centering
\includegraphics{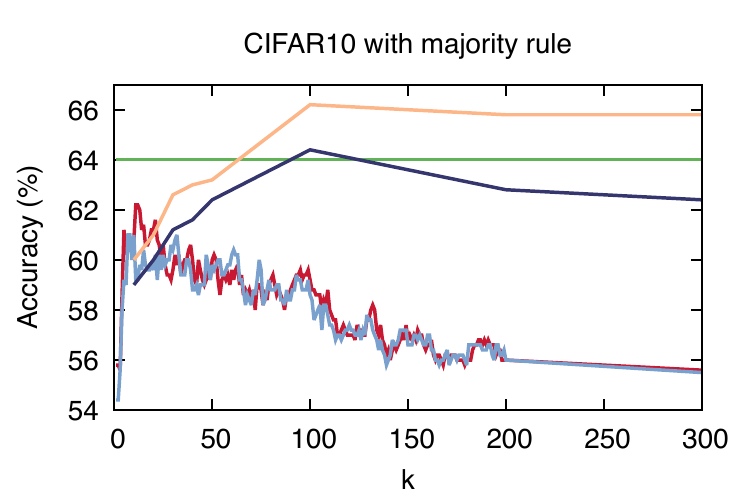}
\caption{CIFAR10 - VGG16}
\label{fig:cif10:maj}
\end{figure}

\begin{figure}[!ht]
\centering
\includegraphics{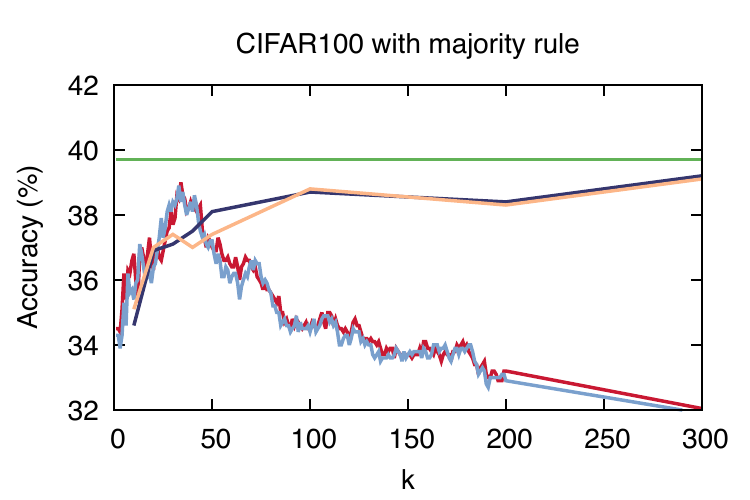}
\caption{CIFAR100 - VGG16}
\label{fig:cif100:maj}
\end{figure}

\begin{figure}[!ht]
\centering
\includegraphics{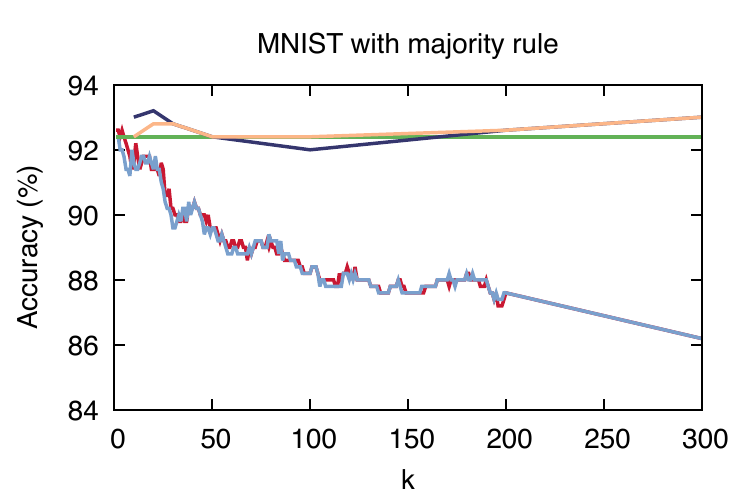}
\caption{MNIST - VGG16}
\label{fig:mnist:maj}
\end{figure}

\begin{figure}[!ht]
\centering
\includegraphics{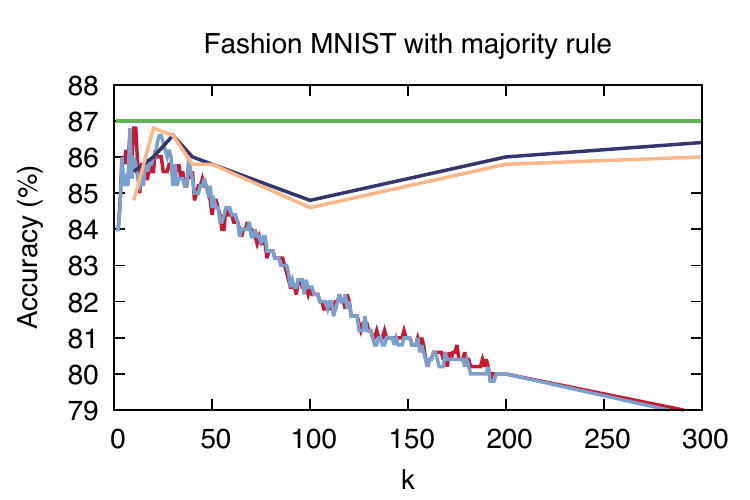}
\caption{Fashion MNIST - VGG16}
\label{fig:fmnist:maj}
\end{figure}

\begin{figure}[!ht]
\centering
\includegraphics{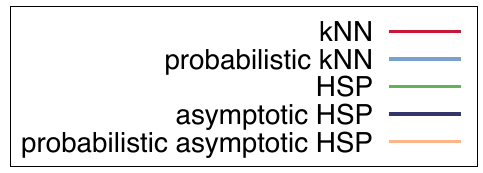}
\caption{Plot legends}
\label{fig:plotlegends}
\end{figure}

We plotted all experiments together to save space. We explain each one of the results appearing in the images, in the order presented in Figure \ref{fig:plotlegends}. Firstly kNN is the vanilla kNN classifier. Probabilistic kNN is the kNN classifier using a probabilistic index. Below we discuss a couple of HSP variants to speed up the computation of the HSP neighborhood of a query. 

\paragraph{Asymptotic and Probabilistic Asymptotic HSP}

The described HSP classifier has linear time complexity, not scaling in large-sized high-dimensional data. One alternative is to compute the HSP inside a ball of a certain radius. A natural radius is the $k$-nearest neighbors.
We call this the {\em Asymptotic HSP}. If we use a probabilistic index to compute the $k$-nearest neighbors of the query, we call the resulting neighborhood {\em Probabilistic Asymptotic HSP}

\section{Experiments}

We tried all values of $k$, from 1 to 300, for the vanilla kNN classifier, the asymptotic HSP, and the probabilistic versions. For the HSP classifier, there is only one value since it is parameter-free. The plots show the accuracy of the classification in the vertical axis. Please notice that any method to select a proper $k$ for kNN will correspond to one value between 1 and 300, dispensing the need to compare to SOTA kNN classifiers. 

\begin{algorithm}
\SetAlgoLined
\KwIn{training samples X and test samples Y}
\KwOut{class labels of Y}
 \For{each $u\in Y$}{
 %$F(u) \leftarrow \varnothing$ \tcp{Set forbidden area}
 $N \leftarrow \varnothing$\;
 $C \leftarrow kNN(X,u,k)$\;
 \While{C is not empty}{
  $v \leftarrow c\in C \mid d(u,c) \leq d(u,c'), \forall c'\in C$\;
  $N.insert(v)$\;
  \For{each $c\in C$}{
  \If{$d(c,u) > d(c,v) $}{
   $C.remove(c)$\;
  }
  }
 }
 $label(u) \leftarrow$ most repeated label in $N$\;
 }
 \caption{Asymptotic probabilistic HSP classifier}
 \label{algorithm:approximatehsp}
\end{algorithm}

Please notice that the probabilistic asymptotic HSP has the same complexity as the probabilistic kNN classifier when using an index like the HNSW \cite{Malkov2020}. The pseudocode is in Algorithm \ref{algorithm:approximatehsp}. 

\begin{figure}[!ht]
\centering
\includegraphics{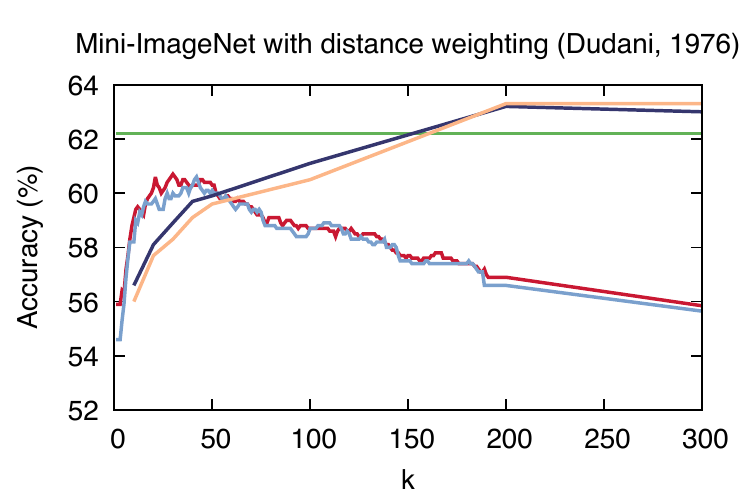}
\caption{Mini-ImageNet - VGG16}
\label{fig:mini:dud}
\end{figure}

\begin{figure}[!ht]
\centering
\includegraphics{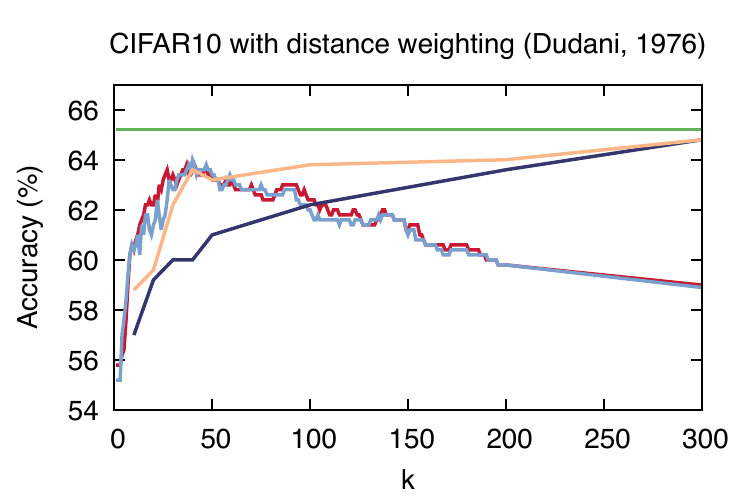}
\caption{CIFAR10 - VGG16}
\label{fig:cif10:dud}
\end{figure}

\begin{figure}[!ht]
\centering
\includegraphics{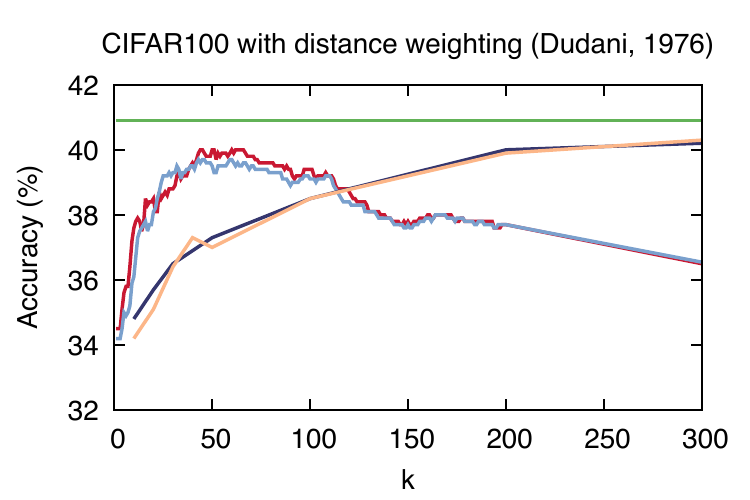}
\caption{CIFAR100 - VGG16}
\label{fig:cif100:dud}
\end{figure}

\begin{figure}[!ht]
\centering
\includegraphics{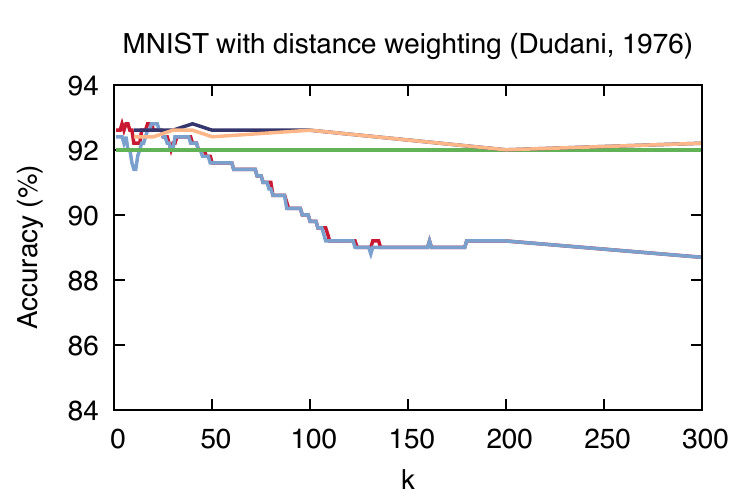}
\caption{MNIST - VGG16}
\label{fig:mnist:dud}
\end{figure}

\begin{figure}[!ht]
\centering
\includegraphics{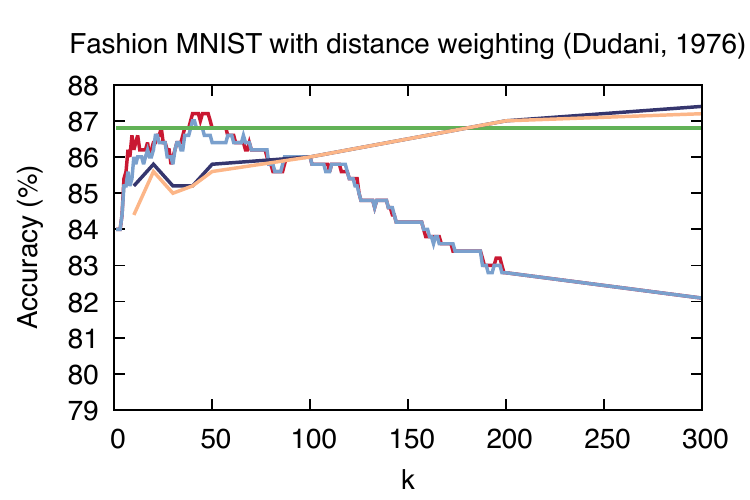}
\caption{Fashion MNIST - VGG16}
\label{fig:fmnist:dud}
\end{figure}

\begin{table}[!ht]
\centering
\footnotesize
\caption{Maximum accuracy percentage for each technique using: 1. Majority rule, 2. Dudani's weighting rule \cite{Dudani1976}, 3. Inverse distance weighting. }
\begin{tabular}{|p{.15cm}|p{1.15cm}|p{0.8cm}|p{0.8cm}|p{.9cm}|p{1cm}|p{1.1cm}|}
\cline{3-7}
\multicolumn{2}{c|}{} & CIFAR \newline 10 & CIFAR \newline 100 & MNIST & Fashion \newline MNIST & Mini\newline ImageNet \\
\hline

\multirow{5}{*}{1} & kNN & 62.2 & 39 & 92.6 & 86.8 & 59.5\\
& P-kNN & 61 & 38.9 & 92.4 & 86.8 & 58.1\\
& HSP & 64 & \textbf{39.7} & 92.4 & 87 & 58.8\\
& A-HSP & 64.4 & \textbf{39.7} & \textbf{93.2} & \textbf{87.2} & \textbf{61.3}\\
& PA-HSP & \textbf{66.4} & \textbf{39.7} & \textbf{93.2} & 87 & 61.1 \\
\hline

\multirow{5}{*}{2} & kNN & 63.8 & 40 & \textbf{92.8} & 87.2 & 60.6\\
& P-kNN & 64 & 39.7 & \textbf{92.8}  & 87 & 60.6\\
& HSP & 65.2 & \textbf{40.9} & 92 & 86.8 & 62.2\\
& A-HSP & 64.8 & \textbf{40.9} & \textbf{92.8}  & \textbf{87.4} & 63.2\\
& PA-HSP & \textbf{67} & \textbf{40.9} & 92.6 & 87.2 & \textbf{63.3} \\
\hline

\multirow{5}{*}{3} & kNN & 62.6 & 38.8 & 92.6 & 86.8 & 59.5\\
& P-kNN & 61.8 & 38.9 & 92.4 & 86.8 & 58.1\\
& HSP & 63.6 & 40.3 & 92.6 & 86.8 & 59.7\\
& A-HSP & 64.8 & \textbf{40.5} & \textbf{93.2} & \textbf{87.4} & \textbf{61.3}\\
& PA-HSP & \textbf{66.8} & \textbf{40.5} & 92.8 & 86.8 & 61\\
\hline

\end{tabular}
\label{table:results}
\end{table}

In the first experiment, we used only the majority rule for the five classifiers. In this case (figures \ref{fig:mini:maj} to \ref{fig:fminist:maj}), the HSP outperforms both versions of kNN in CIFAR 10, Fashion MNIST, and CIFAR 100, being slightly worst than Mini-ImageNet and MNIST. However, the asymptotic versions of HSP outperform kNN in the last two examples. Notice that the asymptotic versions of the HSP have a smoother behavior than kNN, which seems chaotic as a function of $k$.

 It is possible to improve kNN using algorithms giving closer neighbors more influence than further ones. Dudani's work \cite{Dudani1976} is one approach where the weighting function varies with the distance between the query and the considered neighbor in such a manner that the value decreases with increasing the query-to-neighbor distance.
He orders the $k$ neighbors so that $d_{k}$ corresponds to the distance of the furthest neighbor to the query and $d_{1}$ to the nearest one.
 \par 
 
\begin{equation*}
w_{j} = \begin{cases}
\frac{d_{k}-d_{j}}{d_{k}-d_{1}} & d_{k}\neq d_{i}\\
1 & d_{k} = d_{i}
\end{cases}
\end{equation*}

Another fairly general technique is to use inverse distance weighting voting, where the neighbors get to vote on the class of the query with votes weighted by the inverse of their distance to the query.

\begin{equation*}
w_{j} = \frac{1}{d_{j}}
\end{equation*}

In the second experiment, we tried the two versions of distance weighting described above. Due to space restrictions, we only show the plots for Dudani's weighting algorithm. We can observe in figures \ref{fig:mini:dud} to \ref{fig:fmnist:dud} that, similarly to the previous experiment, the HSP outperforms kNN for all datasets, except for MNIST and Fashion MNIST in this case. However, either the asymptotic versions of the HSP or the HSP itself outperform kNN in all cases.\par

The two experiments, plus the inverse distance weighting, are summarized in table \ref{table:results}. In the case of MNIST, with Dudani's rule, kNN matches the performance of our classifier. Our classifier has the additional advantage of being parameter-free or with a smooth behavior in the asymptotic version.

\section{Conclusions and future work}

We presented three new instance-based classifiers, the HSP, Asymptotic HSP, and Probabilistic Asymptotic HSP. We compared our proposal with state of the art kNN classifiers, with optimal parameters (something unachievable in a production environment because there is no ground truth to know the best $k$). Our approach is parameter-free from the point of view of accuracy and a parameter related to the complexity. In all cases, the kNN classifiers at most topped our performance.

With our approach, it is possible to focus on finding a suitable distance function to compare instances when designing a classifier or a data mining task. A parameter-free instance-based classifier could be of help in many applications.

\bibliographystyle{ieeetr}

\begin{thebibliography}{10}

\bibitem{Settouti2016}
N.~Settouti, M.~E.~A. Bechar, and M.~A. Chikh, ``Statistical comparisons of the
  top 10 algorithms in data mining for classification task,'' {\em
  International Journal of Interactive Multimedia and Artificial Intelligence},
  vol.~4, p.~46, 2016.

\bibitem{Wu2008}
X.~Wu, V.~Kumar, Q.~J. Ross, J.~Ghosh, Q.~Yang, H.~Motoda, G.~J. McLachlan,
  A.~Ng, B.~Liu, P.~S. Yu, Z.~H. Zhou, M.~Steinbach, D.~J. Hand, and
  D.~Steinberg, ``Top 10 algorithms in data mining,'' {\em Knowledge and
  Information Systems}, vol.~14, pp.~1--37, 1 2008.

\bibitem{Gou2019}
J.~Gou, H.~Ma, W.~Ou, S.~Zeng, Y.~Rao, and H.~Yang, ``A generalized mean
  distance-based k-nearest neighbor classifier,'' {\em Expert Systems with
  Applications}, vol.~115, pp.~356--372, 1 2019.

\bibitem{Zhang2018}
S.~Zhang, D.~Cheng, Z.~Deng, M.~Zong, and X.~Deng, ``A novel knn algorithm with
  data-driven k parameter computation,'' {\em Pattern Recognition Letters},
  vol.~109, pp.~44--54, 7 2018.

\bibitem{Weinberger2005}
K.~Q. Weinberger, J.~Blitzer, and L.~K. Saul, ``Distance metric learning for
  large margin nearest neighbor classification,'' {\em Advances in neural
  information processing systems}, vol.~18, pp.~1473--1480, 2005.

\bibitem{Xu2013}
Y.~Xu, Q.~Zhu, Z.~Fan, M.~Qiu, Y.~Chen, and H.~Liu, ``Coarse to fine k nearest
  neighbor classifier,'' {\em Pattern Recognition Letters}, vol.~34, 7 2013.

\bibitem{Gallego2018}
A.~J. Gallego, A.~Pertusa, and J.~Calvo-Zaragoza, ``Improving convolutional
  neural networks' accuracy in noisy environments using k-nearest neighbors,''
  {\em Applied Sciences (Switzerland)}, vol.~8, 10 2018.

\bibitem{Zezula2005}
P.~Zezula, G.~Amato, V.~Dohnal, and M.~Batko, {\em Similarity Search: The
  Metric Space Approach}.
\newblock Springer, 2005.

\bibitem{r2018STOC}
A.~Rubinstein, ``Hardness of approximate nearest neighbor search,'' in {\em
  Proceedings of the 50th Annual ACM SIGACT Symposium on Theory of Computing},
  pp.~1260--1268, ACM, 2018.

\bibitem{Malkov2014}
Y.~Malkov, A.~Ponomarenko, A.~Logvinov, and V.~Krylov, ``Approximate nearest
  neighbor algorithm based on navigable small world graphs,'' {\em Information
  Systems}, vol.~45, 9 2014.

\bibitem{Malkov2020}
Y.~A. Malkov and D.~A. Yashunin, ``Efficient and robust approximate nearest
  neighbor search using hierarchical navigable small world graphs,'' {\em IEEE
  Transactions on Pattern Analysis and Machine Intelligence}, vol.~42,
  pp.~824--836, 4 2020.

\bibitem{Li2020}
W.~Li, Y.~Zhang, Y.~Sun, W.~Wang, M.~Li, W.~Zhang, and X.~Lin, ``Approximate
  nearest neighbor search on high dimensional data - experiments, analyses, and
  improvement,'' {\em IEEE Transactions on Knowledge and Data Engineering},
  vol.~32, pp.~1475--1488, 8 2020.

\bibitem{Chavez2006}
E.~Chavez, S.~Dobrev, E.~Kranakis, J.~Opatrny, L.~Stacho, H.~Tejeda, and
  J.~Urrutia, ``Half-space proximal: A new local test for extracting a bounded
  dilation spanner of a unit disk graph,'' {\em International Conference On
  Principles Of Distributed Systems}, pp.~235--245, 2006.

\bibitem{Shi2018}
B.~Shi, L.~Han, and H.~Yan, ``Adaptive clustering algorithm based on knn and
  density,'' {\em Pattern Recognition Letters}, vol.~104, pp.~37--44, 3 2018.

\bibitem{Jiang2007}
L.~Jiang, Z.~Cai, D.~Wang, and S.~Jiang, ``Survey of improving
  k-nearest-neighbor for classification,'' {\em Fourth International Conference
  on Fuzzy Systems and Knowledge Discovery (FSKD 2007)}, vol.~1, pp.~679--683,
  8 2007.

\bibitem{Biswas2018}
N.~Biswas, S.~Chakraborty, S.~S. Mullick, and S.~Das, ``A parameter independent
  fuzzy weighted k-nearest neighbor classifier,'' {\em Pattern Recognition
  Letters}, vol.~101, pp.~80--87, 1 2018.

\bibitem{Tang2020}
B.~Tang, H.~He, and S.~Zhang, ``Mcenn: A variant of extended nearest neighbor
  method for pattern recognition,'' {\em Pattern Recognition Letters},
  vol.~133, pp.~116--122, 5 2020.

\bibitem{Xiang2008}
S.~Xiang, F.~Nie, and C.~Zhang, ``Learning a mahalanobis distance metric for
  data clustering and classification,'' {\em Pattern Recognition}, vol.~41, 12
  2008.

\bibitem{Gautheron2020}
L.~Gautheron, A.~Habrard, E.~Morvant, and M.~Sebban, ``Metric learning from
  imbalanced data with generalization guarantees,'' {\em Pattern Recognition
  Letters}, vol.~133, 5 2020.

\bibitem{Wang2007}
J.~Wang, P.~Neskovic, and L.~N. Cooper, ``Improving nearest neighbor rule with
  a simple adaptive distance measure,'' {\em Pattern Recognition Letters},
  vol.~28, 1 2007.

\bibitem{Li2011}
C.~Li and H.~Li, ``One dependence value difference metric,'' {\em
  Knowledge-Based Systems}, vol.~24, 7 2011.

\bibitem{Li2014}
C.~Li, L.~Jiang, and H.~Li, ``Local value difference metric,'' {\em Pattern
  Recognition Letters}, vol.~49, 11 2014.

\bibitem{Lall1996}
U.~Lall and A.~Sharma, ``A nearest neighbor bootstrap for resampling hydrologic
  time series,'' {\em WATER RESOURCES RESEARCH}, vol.~32, pp.~679--693, 1996.

\bibitem{Zhu2011}
X.~Zhu, S.~Zhang, Z.~Jin, Z.~Zhang, and Z.~Xu, ``Missing value estimation for
  mixed-attribute data sets,'' {\em IEEE Transactions on Knowledge and Data
  Engineering}, vol.~23, pp.~110--121, 2011.

\bibitem{Zhang2017}
S.~Zhang, X.~Li, M.~Zong, X.~Zhu, and D.~Cheng, ``Learning k for knn
  classification,'' {\em ACM Transactions on Intelligent Systems and
  Technology}, vol.~8, 1 2017.

\bibitem{Liu2010}
H.~Liu, S.~Zhang, J.~Zhao, X.~Zhao, and Y.~Mo, ``A new classification algorithm
  using mutual nearest neighbors,'' {\em Proceedings - 9th International
  Conference on Grid and Cloud Computing, GCC 2010}, pp.~52--57, 2010.

\bibitem{Ghosh2006}
A.~K. Ghosh, ``On optimum choice of k in nearest neighbor classification,''
  {\em Computational Statistics and Data Analysis}, vol.~50, 7 2006.

\bibitem{ktree2018}
S.~Zhang, X.~Li, M.~Zong, X.~Zhu, and R.~Wang, ``Efficient knn classification
  with different numbers of nearest neighbors,'' {\em IEEE Transactions on
  Neural Networks and Learning Systems}, vol.~29, pp.~1774--1785, 5 2018.

\bibitem{Jaromczyk1992}
J.~W. Jaromczyk and G.~T. Toussaint, ``Relative neighborhood graphs and their
  relatives,'' {\em Proceedings of the IEEE}, vol.~80, pp.~1502--1517, 1992.

\bibitem{corral}
R.~Corral-Corral, E.~Chavez, and G.~Del~Rio, ``Machine learnable fold space
  representation based on residue cluster classes,'' {\em Computational Biology
  and Chemistry}, vol.~59, pp.~1--7, 2015.

\bibitem{aguilera}
L.~Aguilera-Mendoza, Y.~Marrero-Ponce, C.~R. Garc{\'\i}a-Jacas, E.~Chavez,
  J.~A. Beltran, H.~A. Guillen-Ramirez, and C.~A. Brizuela, ``Automatic
  construction of molecular similarity networks for visual graph mining in
  chemical space of bioactive peptides: an unsupervised learning approach,''
  {\em Scientific reports}, vol.~10, no.~1, pp.~1--23, 2020.

\bibitem{Deng2009}
J.~Deng, W.~Dong, R.~Socher, L.-J. Li, K.~Li, and L.~Fei-Fei, ``Imagenet: A
  large-scale hierarchical image database,'' {\em 2009 IEEE Conference on
  Computer Vision and Pattern Recognition}, pp.~248--255, 6 2009.

\bibitem{Wang2019}
J.~Wang, Y.~Chen, S.~Hao, X.~Peng, and L.~Hu, ``Deep learning for sensor-based
  activity recognition: A survey,'' {\em Pattern Recognition Letters},
  vol.~119, 3 2019.

\bibitem{LeCun2010}
Y.~LeCun and C.~Cortes, ``Mnist handwritten digit database,'' 2010.

\bibitem{Xiao2017}
H.~Xiao, K.~Rasul, and R.~Vollgraf, ``Fashion-mnist: a novel image dataset for
  benchmarking machine learning algorithms,'' 8 2017.

\bibitem{Krizhevsky2009}
A.~Krizhevsky, ``Learning multiple layers of features from tiny images,'' 2009.

\bibitem{Vinyals2016}
O.~Vinyals, G.~Deepmind, C.~Blundell, T.~Lillicrap, K.~Kavukcuoglu, and
  D.~Wierstra, ``Matching networks for one shot learning,'' 2016.

\bibitem{Dudani1976}
S.~A. Dudani, ``The distance-weighted k-nearest-neighbor rule let each pattern
  pi in the training set (collection of correctly,'' 1976.

\end{thebibliography}

\end{document}